\documentclass[sn-mathphys-num]{sn-jnl}% Math and Physical Sciences Numbered Reference Style 
\usepackage[inkscapelatex=false]{svg}
\usepackage{caption}
\usepackage{graphicx}%
\usepackage{multirow}%
\usepackage{amsmath,amssymb,amsfonts}%
\usepackage{amsthm}%

\usepackage{mathrsfs}%
\usepackage[title]{appendix}%
\usepackage{xcolor}%
\usepackage{textcomp}%
\usepackage{manyfoot}%
\usepackage{booktabs}%
\usepackage{algorithm}%
\usepackage{algorithmicx}%
\usepackage{algpseudocode}%
\usepackage{listings}%

\usepackage[numbers]{natbib}%

\newtheoremstyle{thmstyleone}{}{}{}{}{}{}{\newline}{\thmname{#1}\thmnumber{ #2}\thmnote{ (#3)}}
\theoremstyle{thmstyleone}%
\newtheoremstyle{thmstyletwo}{}{}{}{}{}{}{\newline}{\thmname{#1}\thmnumber{ #2}\thmnote{ (#3)}}
\theoremstyle{thmstyletwo}%

\newtheoremstyle{thmstylethree}{}{}{}{}{}{}{\newline}{\thmname{#1}\thmnumber{ #2}\thmnote{ (#3)}}
\theoremstyle{thmstylethree}%

\begin{document}

\title[Article Title]{AD-GPT: Large Language Models in Alzheimer's Disease}


\author[1]{\fnm{Ziyu} \sur{Liu}}\email{zl23565@uga.edu}

\author[2]{\fnm{Lintao} \sur{Tang}}\email{lt20ca@fsu.edu}

\author[3]{\fnm{Zeliang} \sur{Sun}}\email{zs58484@uga.edu}

\author[4]{\fnm{Zhengliang} \sur{Liu}}\email{zl18864@uga.edu}

% \author[5]{\fnm{Tianyang} \sur{Zhong}}\email{tzhong3@ualberta.ca}

\author[5]{\fnm{Yanjun} \sur{Lyu}}\email{yxl9168@mavs.uta.edu}

\author[4]{\fnm{Wei} \sur{Ruan}}\email{wei.ruan@uga.edu}

\author[2]{\fnm{Yangshuang} \sur{Xu}}\email{yx23e@fsu.edu}

\author[2]{\fnm{Liang} \sur{Shan}}\email{ls21s@fsu.edu}

\author[1]{\fnm{Jiyoon} \sur{Shin}}\email{js12571@uga.edu}

\author[2]{\fnm{Xiaohe} \sur{Chen}}\email{xchen8@fsu.edu}

\author[5]{\fnm{Dajiang} \sur{Zhu}}\email{dajiang.zhu@uta.edu}

\author[4]{\fnm{Tianming} \sur{Liu}}\email{tliu@uga.edu}

\author*[1]{\fnm{Rongjie} \sur{Liu}}\email{rjliu@uga.edu}

\author*[3]{\fnm{Chao} \sur{Huang}}\email{chaohuang@uga.edu}

\affil[1]{\orgdiv{Department of Statistics}, \orgname{University of Georgia}, \orgaddress{\city{Athens}, \state{GA}, \country{USA}}}

\affil[2]{\orgdiv{Department of Statistics}, \orgname{Florida State University}, \orgaddress{\city{Tallahassee}, \state{FL}, \country{USA}}}

\affil[3]{\orgdiv{Department of Epidemiology \& Biostatistics}, \orgname{University of Georgia}, \orgaddress{\city{Athens}, \state{GA}, \country{USA}}}

\affil[4]{\orgdiv{School of Computing}, \orgname{University of Georgia}, \orgaddress{\city{Athens}, \state{GA}, \country{USA}}}

% \affil[5]{\orgdiv{Department of Mathematical and Statistical Science}, \orgname{University of Alberta}, \orgaddress{\city{Edmonton},\country{Canada}}}

\affil[5]{\orgdiv{Department of Computer Science and Engineering}, \orgname{University of Texas at Arlington}, \orgaddress{\city{Arlington}, \state{TX}, \country{USA}}}


\abstract{Large language models (LLMs) have emerged as powerful tools for medical information retrieval, yet their accuracy and depth remain limited in specialized domains such as Alzheimer's disease (AD), a growing global health challenge. To address this gap, we introduce AD-GPT, a domain-specific generative pre-trained transformer designed to enhance the retrieval and analysis of AD-related genetic and neurobiological information. AD-GPT integrates diverse biomedical data sources, including potential AD-associated genes, molecular genetic information, and key gene variants linked to brain regions. We develop a stacked LLM architecture combining Llama3 and BERT, optimized for four critical tasks in AD research: (1) genetic information retrieval, (2) gene–brain region relationship assessment, (3) gene–AD relationship analysis, and (4) brain region–AD relationship mapping. Comparative evaluations against state-of-the-art LLMs demonstrate AD-GPT’s superior precision and reliability across these tasks, underscoring its potential as a robust and specialized AI tool for advancing AD research and biomarker discovery.}

\maketitle
\section{Main}\label{sec1}
% \textcolor{red}{For the Introduction, we have four components: 1. AD and AD-related information retrieval; 2. LLM and its applications in AD and AD-related information retrieval; 3. Challenges in existing LLMs; 4. Our contributions.}\\~\\
% \textcolor{blue}{1 and 2 assigned to Yangshuang (1 \& 2 finished), 3 and 4 assigned to Ziyu. (3 \& 4 are finished. Let me know if anything needs to be polished -Ziyu 1-15-2025 }\\~\\

% introduce AD and the challenge in integrating findings from literature
Alzheimer’s disease (AD) is a progressive neurodegenerative disorder that profoundly impacts memory, cognition, and behavior \cite{scheltens2021alzheimer}. It typically begins with subtle memory impairment and confusion, gradually advancing to severe deficits in language, spatial orientation, and executive function \cite{weiner2008language}. 
%As the disease advances, patients may also experience changes in personality, mood swings, anxiety, depression, and delusions \cite{terracciano2019personality}. Eventually, AD leads to severe loss of independence, requiring full-time care. 
% Diagnosis of AD involves a comprehensive evaluation, including clinical evaluations and neuroimaging techniques, for example, the magnetic resonance imaging \cite{deture2019neuropathological}. 
As the global burden of AD continues to rise, large-scale biomedical studies have emerged, generating vast datasets across diverse modalities, including neuroimaging, genomics, neurocognitive assessments, and clinical profiles \cite{gao2023chinese}. These extensive datasets have facilitated the identification of key biomarkers implicated in AD onset and progression, offering critical insights into its underlying pathological mechanisms. However, efficiently integrating these findings from existing literature and databases still remains a significant challenge, which underscores the immense value offered by large-scale studies in driving the advancement of diagnostic and therapeutic strategies for AD patients.

% introduce IR and its importance in AD
Information retrieval (IR) is an important tool that focuses on the identification and extraction of relevant information from vast datasets or document collections \cite{dash2021biomedical}. In the context of AD research, IR plays a pivotal role by enabling the efficient access to critical data, thereby supporting a wide array of research applications. For instance, IR facilitates the retrieval of biomarker data, such as beta-amyloid plaque levels or tau protein concentrations, which provide valuable insights into disease mechanisms and enhance early diagnostic capabilities \cite
{rowan2012information}. Additionally, IR is instrumental in accessing medical imaging data from repositories, allowing researchers to track and analyze patterns of brain atrophy and functional changes over time \cite
{trojachanec2017longitudinal}. Furthermore, IR aids in identifying genetic studies \cite{shatkay2002information}, which could include those investigating APOE polymorphisms, which are strongly linked to AD risk. Therefore, IR is essential for navigating the growing body of AD research, ensuring that crucial data is accessible for advancing our understanding and treatment of AD.

% introduce LLM and LLm related IR, 
Large Language Models (LLMs) are advanced tools in natural language processing (NLP) that have demonstrated remarkable capabilities across various domains. These models, such as the GPT (Generative Pre-trained Transformer) series,
including ChatGPT and Llama, 
%, and recently developed DeepSeek \cite{bi2024deepseek}
are designed to understand and generate human-like text, which makes them particularly effective for addressing challenges in IR \cite{lu2024llm}. In particular, LLM-based IR can streamline the process of retrieving complex biomedical databases, such as clinical records that focus on the relationships between various phenotypes and genotypes, making it an invaluable tool for AD researchers
\cite{dai2024bias}.

Although LLMs have demonstrated broad applicability in IR within the medical domain \cite{dai2023ad,chen2023meditron, wang2023huatuo, li2023chatdoctor}, they also exhibit several critical limitations. One of the most significant concerns is their propensity for hallucination, a phenomenon in which models generate responses that appear confident yet are factually incorrect or nonsensical \cite{ji2023survey}. This issue becomes particularly pronounced in tasks requiring domain-specific expertise, such as medical and legal inquiries, where LLMs frequently produce fabricated information presented with unwarranted certainty \cite{pugliese2024accuracy}. 
%A research \cite{pugliese2024accuracy} evaluated ChatGPT’s performance in addressing questions related to nonalcoholic fatty liver disease, and they highlighted a case in which
%the answer generated by ChatGPT regarding the use of herbal remedies in nonalcoholic fatty liver disease was considered potentially
%harmful and incorrect by the majority of the key opinion leaders. In tasks to analyze 2400 real patient cases of four common abdominal pathologies (appendicitis, pancreatitis, cholecystitis, diverticulitis), current LLMs perform significantly worse than clinicians\cite{hager2024evaluating}.
%In clinical contexts, where decisions can have profound implications for patient well-being, the consequences of hallucination can range from being mildly misleading to outright catastrophic. Erroneous advice or information provided to patients may lead to adverse health outcomes, particularly if such information is acted upon without consulting a healthcare professional \cite{giuffre2024evaluating}.
Beyond hallucination, LLMs often fall short in providing depth and comprehensiveness, particularly in specialized contexts. While models like ChatGPT can generate largely accurate responses, they are frequently criticized for their lack of nuanced understanding. For instance, in the field of epilepsy, LLM-generated content is often superficial, failing to capture the intricacies of the condition \cite{hristidis2023chatgpt}. Similarly, in genetics, empirical evaluations suggest that ChatGPT’s performance is comparable to that of human respondents, offering no clear advantage in accuracy or insight \cite{duong2024analysis}. These limitations largely stem from the models’ constrained exposure to medical-domain knowledge and their inherent difficulty in navigating the complexities of clinical reasoning.

To address these limitations, researchers have sought to enhance LLMs by fine-tuning them with domain-specific corpora. Notable efforts, such as  Meditron \cite{chen2023meditron}, HuaTuo \cite{wang2023huatuo}, and ChatDoctor \cite{li2023chatdoctor}, exemplify attempts to embed biomedical expertise within LLM architectures, improving their applicability in clinical and research settings. Meanwhile, given the expansive body of AD research spanning genomics, proteomics, and other disciplines, there exists a wealth of literature and publicly available datasets. Despite these advances, to date, only a single research group has proposed an LLM-driven IR system tailored specifically for AD \cite{dai2023ad}. However, this model primarily functions as a tool for retrieving news updates and extracting spatio-temporal data, rather than leveraging rigorously curated datasets or generating domain-informed insights. This reliance on unstructured, non-validated sources raises concerns regarding information reliability, data validity, and the model’s constrained analytical depth. Therefore, bridging this gap requires the development of specialized LLM frameworks capable of harmonizing structured biomedical databases with unstructured textual knowledge, ultimately unlocking their full potential for advancing AD research.

\begin{figure}[h!]
    \centering
    \includegraphics[width=\textwidth]{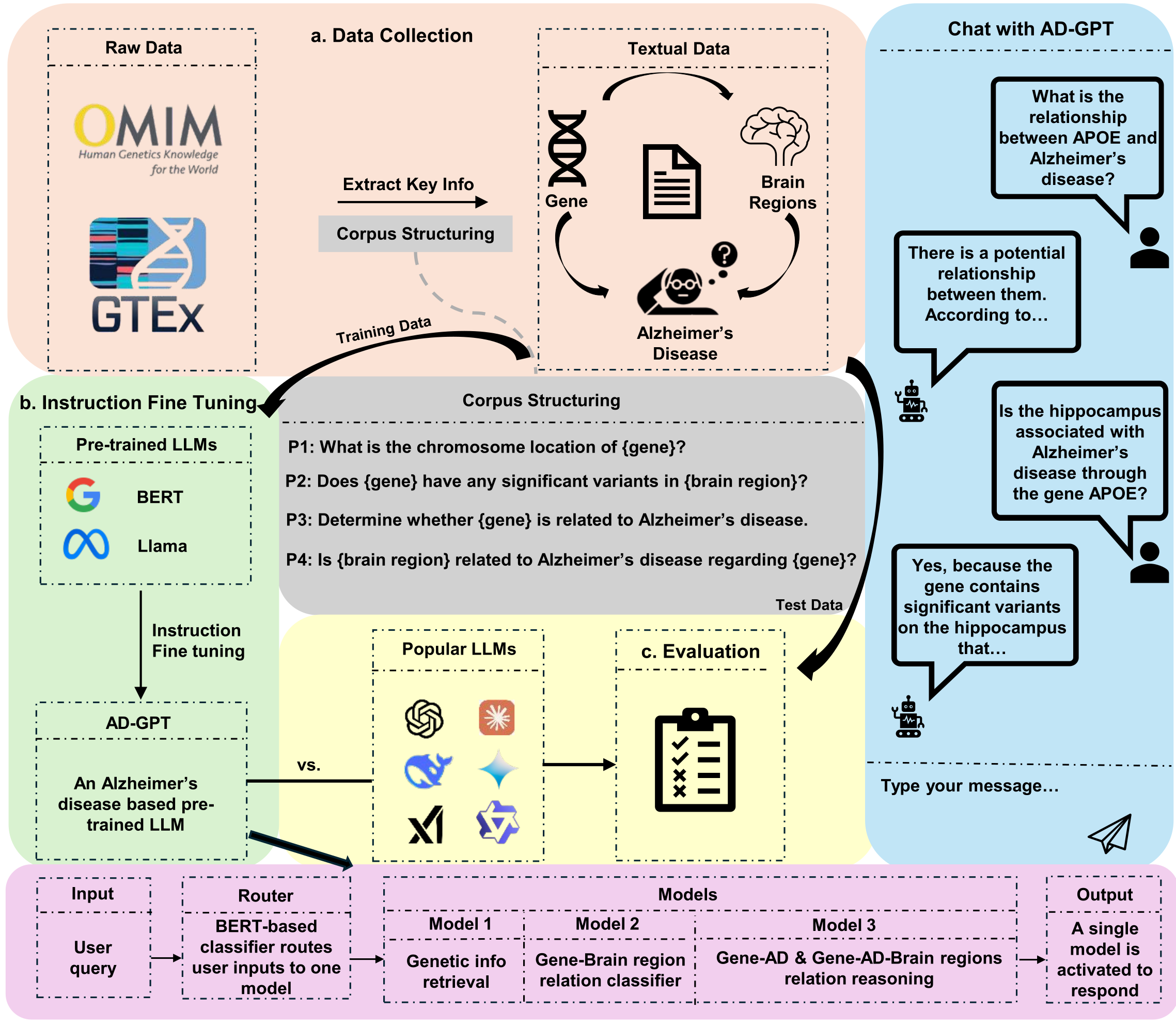} % Adjust the filename if needed
    \caption{{\bf Overview of the AD-GPT workflow.} AD-GPT integrates multimodal data from publicly available databases, including OMIM and the GTEx Project, to systematically extract AD-related information. This information is curated into four specialized textual corpora, each designed to support distinct research tasks. The system builds upon pre-trained Llama and BERT models, further fine-tuned to enhance performance in domain-specific applications. Model efficacy is benchmarked against state-of-the-art language models using diverse evaluation metrics. To facilitate accessibility and usability, the entire framework is encapsulated within a Docker container and equipped with an interactive GUI, enabling seamless deployment for AD research.}
    \label{fig:pipeline}
\end{figure}

In this paper, we introduced AD-GPT, a fine-tuned stacked model designed to systematically integrate domain-specific LLMs to enhance AD research (Fig. \ref{fig:pipeline}). Our approach establishes a structured workflow that begins with the acquisition and curation of high-quality genetic and transcriptomic datasets from reputable public sources, such as the Online Mendelian Inheritance in Man (OMIM) \cite{omim} and the Genotype-Tissue Expression (GTEx) Project \cite{GTEx}. 
%provide a robust foundation for constructing an AD-specific LLM-based IR. 
These data were meticulously processed to construct a specialized textual corpus for supervised fine-tuning, enabling the model to capture the intricate relationships between genetic factors, brain regions, and AD pathology. 
Beginning with comprehensive data acquisition from reputable public genetic data sources, we structured genetic, transcriptomics, and AD-related data to build the textual corpus for supervised fine-tuning. Four distinct corpora were constructed, each tailored specifically for one of four defined tasks: genetic information retrieval (Task 1), gene-brain region relationship assessment (Task 2), gene-AD relationship analysis (Task 3), and brain region-AD relationship mapping (Task 4). To effectively handle this multi-task problem, we implemented a stacked model where a BERT classifier deterministically classifies user queries and selects the appropriate task model to generate responses. 
BERT was utilized for Task 2 due to its effectiveness in classification scenarios, while the Llama model was employed for the remaining tasks, which required more nuanced generative reasoning capabilities. This structured approach ensures precise, contextually relevant responses to various AD-related inquiries.
Furthermore, to rigorously assess AD-GPT’s capabilities, we conducted extensive comparative evaluations against state-of-the-art LLMs across the four AD-related tasks. The results highlight AD-GPT’s superior performance, underscoring its potential as a reliable and precise tool for advancing AD research.

To ensure seamless deployment and user-friendly interaction, AD-GPT is encapsulated within a Docker container, integrating all model components and the FastAPI backend into a self-contained environment. This streamlined architecture minimizes configuration requirements, allowing for effortless system initialization. The platform features an HTML-based graphical user interface (GUI), providing an intuitive and accessible framework for researchers and clinicians to engage with the model. Moreover, the modular design of AD-GPT facilitates scalability and iterative enhancements, enabling seamless integration of future updates and refinements while maintaining system stability and efficiency.

\section{Results}\label{sec2}

To evaluate the performance of AD-GPT, we benchmarked it against several state-of-the-art LLMs, including ChatGPT o1 \cite{openai2024learning}, Claude series including Claude3.5-Haiku and Claude3.7-Sonnet \cite{anthropic_claude_3.5_haiku}, DeepSeek-R1-Distill-Llama-8B \cite{deepseek2025}, Gemini1.5-Flash \cite{team2024gemini}, Llama series including Llama2, Llama3.1, and Llama3.2 \cite{meta2025Llama31}, Qwen2.5 \cite{qwen2.5}, and Grok3 \cite{xai2025grok3}. These models vary in scale and optimization strategies, providing a diverse benchmark for evaluating biomedical knowledge retrieval and reasoning capabilities (Method Section \ref{sec:LLM}). 

For Tasks 1 and 2, which involved structured queries and binary classification, we utilized standard evaluation metrics, including accuracy, precision, recall, and F1-score. The datasets for these tasks comprised 2,160 and 10,140 instruction-output pairs, respectively, covering diverse combinations of genes, brain regions, and query types (Method Section \ref{sec:data_collection} and \ref{sec:corpora}). A randomly selected 10\% subset of each dataset was designated as the test set for computing performance metrics.

In contrast, Tasks 3 and 4 required complex text generation and advanced reasoning capabilities, necessitating qualitative evaluation via expert assessment. We enlisted multiple domain experts to systematically evaluate the responses generated by AD-GPT and comparator LLMs. The assessment criteria were based on two key metrics: relevance, which measured the contextual appropriateness of responses, and precision, which quantified the factual correctness of generated content relative to established references. Each response was rated on a 0–5 scale, with higher scores indicating superior performance.
To ensure robustness and fairness in evaluation, we generated 20 novel queries for each of Tasks 3 and 4, all of which were unseen during model fine-tuning. Responses from AD-GPT and benchmark models were independently rated by three domain experts. The final performance scores were determined by averaging the ratings across experts, providing an unbiased and consistent assessment of model capabilities.
%Below, we summarize the key characteristics of each model.\\~\\
% \textcolor{red}{Fo each table, we should compare with more LLM models, including Llama, ChatGPT, Claude, Gemini. In particular, different versions of each model should be involved as well.} \\~\\
% \textcolor{blue}{Lintao}\\

\begin{figure}[h!]
    \centering
    \includegraphics[width=\textwidth]{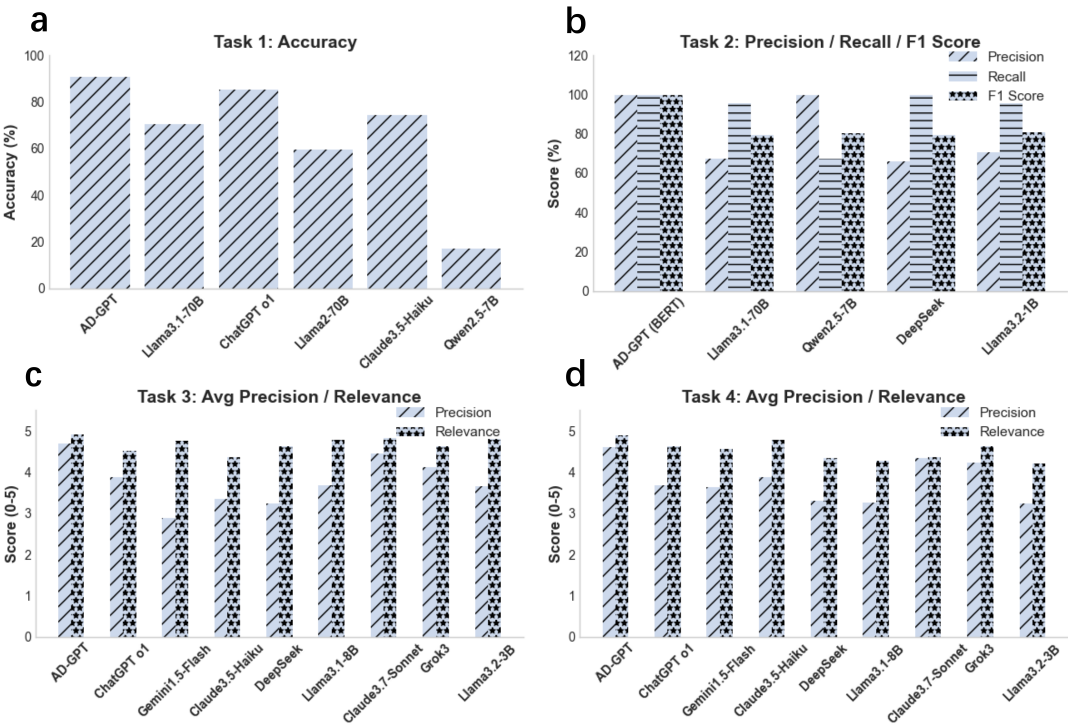}
    \caption{{\bf Performance comparison of different language models across four tasks.}
(a) Accuracy in Task 1 for answering gene-attribute-related questions.
(b) Precision, recall, and F1 score in Task 2 for identifying gene–brain region relationships.
(c) Average precision and relevance scores (0–5) in Task 3 rated by experts.
(d) Average precision and relevance scores (0–5) in Task 4 rated by experts.}
    \label{fig:4tasksresult}
\end{figure}

In Task 1, our base AD-GPT outperformed all competing models across all evaluated metrics (Fig. \ref{fig:4tasksresult} (a) and Supplementary Table 1). Notably, AD-GPT achieved an accuracy of 90.84\%, substantially surpassing Qwen2.5 (17.14\%), Llama2 (59.59\%), Llama3.1-70B (70.52\%) and Claude3.5-Haiku (74.28\%), and demonstrating a clear advantage over ChatGPT o1 (85.33\%).  
%This performance underscores the model’s robustness in accurately resolving gene attribute-related queries. 
While ChatGPT o1 performed well with structured inputs, its accuracy declined for shorter or ambiguous queries, whereas Llama3.2-1B struggled markedly with less well-defined input structures. These results highlight the critical role of domain-specific optimization, as our fine-tuned AD-GPT consistently delivered superior accuracy and reliability, establishing it as a powerful tool for gene attribute-related question answering.

% \begin{table}[ht]
%     \captionsetup{position=above}
%     \caption{\small Performance comparison of models for Task 1. The metrics reflect each model's ability to answer gene-related questions accurately and consistently.}
%     \centering
%     \renewcommand{\arraystretch}{1.1} % Adjust row spacing
%     \setlength{\tabcolsep}{4pt} % Reduce column spacing
%     \small % Reduce font size
%     \begin{tabular}{lcccccc}
%         \toprule
%         \textbf{Model} &\textbf{Params} &\textbf{Accuracy (\%)}\\
%         \midrule
%         Base AD-GPT  &8B   & 90.84 \\
%         Llama 3.1   &70B   & 70.52 \\
%         ChatGPT o1  & $\approx$ 200B    & 85.33 \\
%         Llama 2 &70B      & 59.59 \\
%         Claude3.5-Haiku & $\approx$ 175B      & 74.28 \\
%         Qwen2.5& 7B & 17.14\\
%         \bottomrule
%     \end{tabular}
    
%     \label{tab:task1_performance}
% \end{table}
% \begin{table}[ht]
%     \centering
%     \begin{tabular}{|l|c|c|c|c|}
%         \hline
%         \textbf{Model} &\textbf{Params} &\textbf{Accuracy (\%)}\\
%         \hline
%         Base AD-GPT  &8B   & 90.84 \\
%         Llama 3.1   &70B   & 70.52 \\
%         ChatGPT o1  & $\approx$ 200B    & 85.33 \\
%         Llama 2 &70B      & 59.59 \\
%         Claude3.5-Haiku & $\approx$ 175B      & 74.28 \\
%         Qwen2.5& 7B & 17.14\\
%         \hline
%     \end{tabular}
%     \caption{Performance comparison of models for Task 1. The metrics reflect each model's ability to answer gene-related questions accurately and consist1ently.}
%     \label{tab:task1_performance}
% \end{table}
The performance results for Task 2 highlight the exceptional capability of our BERT-based AD-GPT model in evaluating gene–brain region relationships (Fig. \ref{fig:4tasksresult} (b) and Supplementary Table 1). Notably, our model achieved a perfect score of 100\% across accuracy, precision, recall, and F1-score, demonstrating its robustness in identifying significant variants that influence gene expression or splicing regulation within specific brain regions. In contrast, alternative models, including DeepSeek Distilled Llama3.1-8B, Llama3.2-1B, and Llama3.1-70B, exhibit substantially lower accuracies (approximately 66–70\%). While Qwen2.5-7B achieves a comparable accuracy of 99.83\%, its reduced recall and F1-score suggest inconsistencies in performance. These findings underscore the effectiveness of our domain-specific fine-tuning in delivering precise and reliable analyses, establishing AD-GPT as a powerful tool for evaluating gene–brain region relationships.

For Tasks 3 and 4, we benchmarked our fine-tuned model against state-of-the-art LLMs, including Llama3.1-8B, Llama3.2-3B, Claude3.5-Haiku, Claude3.7-Sonnet, Gemini1.5-Flash, Grok3, DeepSeek-R1-Distill-Llama-8B, and ChatGPT o1. 
%To assess model performance, domain experts in gene biology evaluated responses using two key metrics: relevance, which measures the contextual appropriateness of a response to the query, and precision, which reflects the factual correctness of the generated content relative to a reference. Both metrics were scored on a scale of 0 to 5, with higher scores indicating superior performance.
AD-GPT consistently outperformed all competing models across both tasks (Fig. \ref{fig:4tasksresult} (c)-(d) and Supplementary Table 2). In Task 3, AD-GPT achieved the highest scores, with a precision of 4.70 and a relevance of 4.92, whereas alternative models, including DeepSeek-R1-Distill-Llama-8B, ChatGPT o1, and Gemini1.5-Flash, exhibited lower precision. Similarly, in Task 4, AD-GPT maintained its lead, scoring 4.60 in precision and 4.90 in relevance, surpassing Llama models, DeepSeek-R1-Distill-Llama-8B, Gemini1.5-Flash, ChatGPT o1, and Claude3.5-Haiku. These findings highlight a critical limitation of general-purpose models: while some exhibit reasonable relevance, they frequently struggle with factual accuracy and reliable citation. The superior performance of AD-GPT underscores the efficacy of domain-specific fine-tuning in enhancing both precision and contextual relevance in specialized applications.

\begin{figure}[h!]
    \centering
    \includegraphics[width=\textwidth]{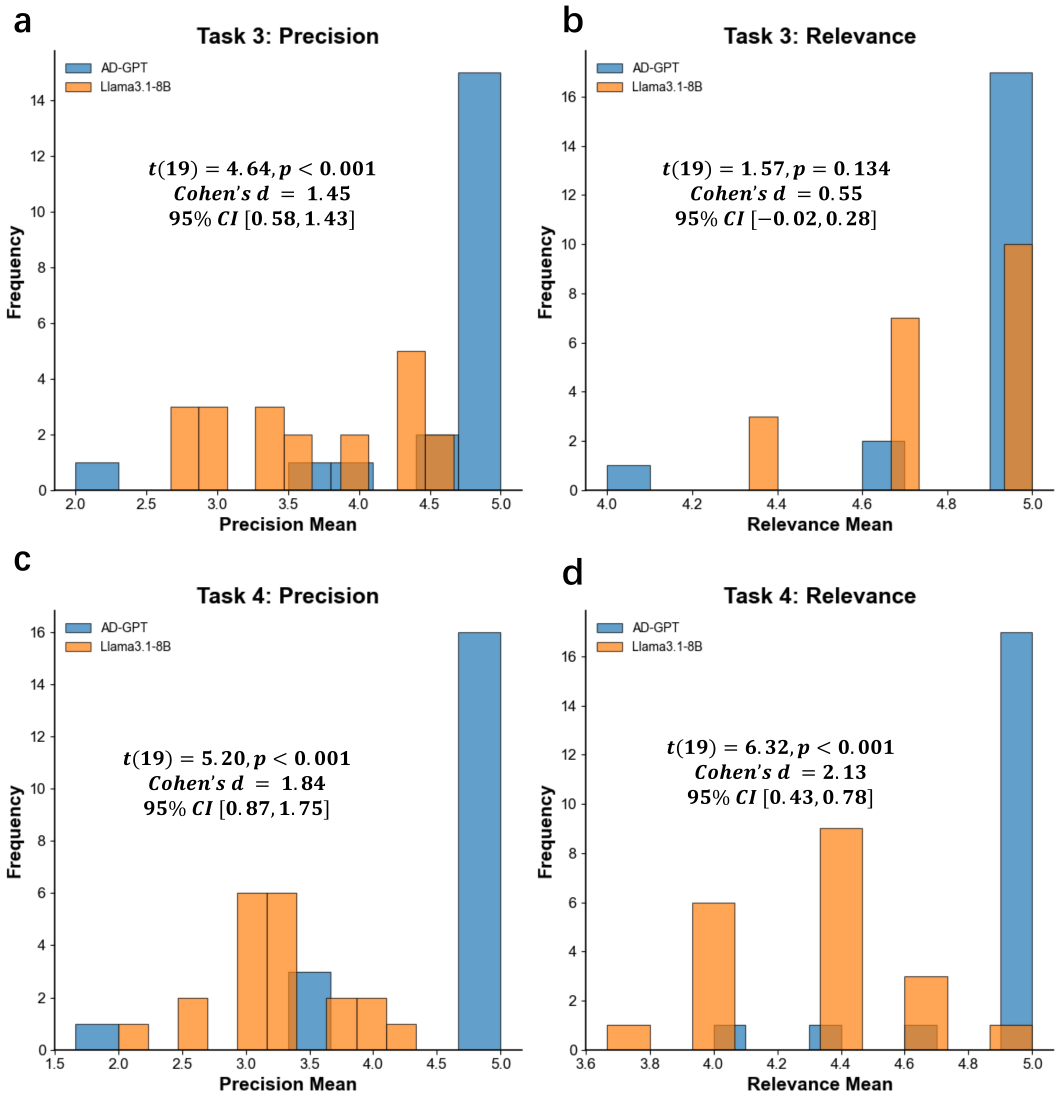}
    \caption{{\bf Performance improvements of AD-GPT following QLoRA fine-tuning on Tasks 3 and 4.}
The QLoRA fine-tuning process substantially shifted the distributions of precision and relevance scores for both tasks. Two-sided paired $t$-tests were conducted to evaluate the statistical significance of these metric improvements. Cohen’s d and 95\% confidence intervals are reported to quantify the effect sizes.}
    \label{fig:finetune_task34}
\end{figure}

To comprehensively evaluate the effectiveness of fine-tuning, we compared the performance of the fine-tuned model AD-GPT against its pre-trained counterpart, Llama3.1-8B, on Tasks 3 and 4. AD-GPT was fine-tuned using Quantized Low-Rank Adaptation (QLoRA), specifically targeting the grouped-query attention (GQA) and feed-forward layers.  LoRA introduced about $134$ million new weights, which is 1.675\% of the total number of weights in Llama3.1-8B.
This fine-tuning strategy substantially improved complex text generation and advanced reasoning, where expert evaluations revealed a marked increase in both relevance and precision, i.e., 20.6\% and 2.8\% in Task 3 while 27\% and 12.4\% in Task 4 (Supplementary Table 2). Furthermore,  as shown in Fig. \ref{fig:finetune_task34}, LoRA fine-tuning dramatically shifted the distribution of precision scores for Task 3 ($t(19)$=4.64, $P<$0.001, Cohen's d=1.45, 95\% CI 0.58-1.43; two sided). In Task 4, AD-GPT further demonstrated superior performance in both precision ($t(19)$=5.20, $P<$0.001, Cohen's d=1.84, 95\% CI 0.87-1.75; two sided) and relevance ($t(19)$=6.32, $P<$0.001, Cohen's d=2.13, 95\% CI 0.43-0.78; two sided) ratings, underscoring the effectiveness of QLoRA fine-tuning in enhancing model capability for AD-related biomedical tasks.

\begin{figure}[h!]
    \centering
    \includegraphics[width=\textwidth]{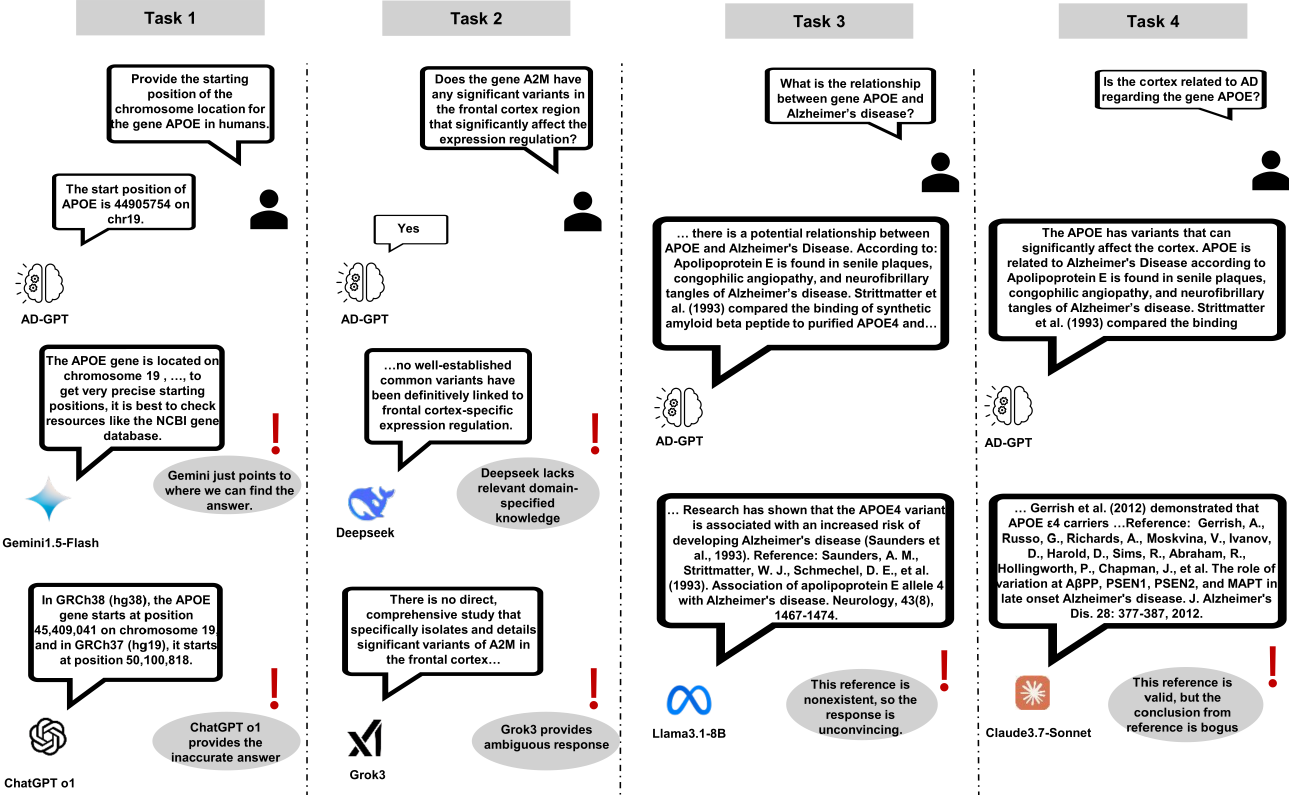}
    \caption{\textbf{Representative questions randomly selected from Tasks 1-4 and the responses generated by AD-GPT and its counterparts. }AD-GPT consistently demonstrates more accurate, specific, and evidence-backed answers across a range of genomic and disease-association queries.}
    \label{fig:limitations}
\end{figure}

To further evaluate the reliability of AD-GPT and competing LLMs, we randomly selected one representative question from each task and systematically analyzed the responses generated by AD-GPT and its counterparts (Fig. \ref{fig:limitations}). AD-GPT consistently provided accurate and contextually appropriate answers across all tasks. In contrast, we identified notable deficiencies in competing LLMs, which varied depending on the nature of the task. In Tasks 1 and 2, several models, including ChatGPT o1, Grok3, Gemini1.5-Flash, and DeepSeek-R1-Distill-Llama-8, either failed to generate responses or produced inaccurate answers, likely due to their limited exposure to domain-specific knowledge. In Tasks 3 and 4, other models, such as Llama3.1-8B and Claude3.7-Sonnet, frequently generated ambiguous statements or fabricated references to support their conclusions, undermining their credibility. These findings highlight the importance of domain adaptation in LLM development and underscore the advantages of AD-GPT in handling specialized inquiries with greater precision and reliability.

% , our approach ensures accuracy, transparency, and efficiency. The model's self-contained nature allows for secure, on-premise deployment, making it ideal for medical and research applications where data privacy is paramount. With a highly curated AD-specific database, our stacked architecture provides a robust, low-latency solution tailored for bioinformatics and neuroscience research.

\section{Discussion}
 % This image highlights key issues with LLM-generated responses in biomedical research. Gemini provides an incomplete answer by directing users to external sources without specific details, demonstrating indecisiveness. Llama, on the other hand, fabricates a reference, leading to false credibility and unreliable scientific claims. These limitations emphasize the need for domain-specific fine-tuning in biomedical AI applications.
%\subsection{Model and Metric Selection for Task Comparisons}\label{subsec4}
%Frist, we briefly discuss the models and metrics used for performance comparison.  
Model and metric selection for performance comparison have already been carried out.
In Task 1, we restricted our comparison to LLMs with at least some genetic knowledge. Smaller models, such as Llama3.1-7B, Gemini1.5-Flash, and DeepSeek-R1-Distilled-Llama-3-7B, struggled with these queries, frequently producing indecisive or incorrect responses.
% (see Figure~\ref{fig:limitations})
In Task 2, given that our model is based on BERT and relatively lightweight, we conducted a horizontal comparison against similarly compact architectures, including Llama3.2-1B and Llama3.1-8 B. As Task 2 is formulated as a binary classification problem, we employed accuracy, precision, recall, and F1 score as evaluation metrics to ensure a rigorous and comprehensive assessment.
For all tasks except Task 2, we selected ChatGPT o1 as an alternative baseline but excluded models such as GPT-3 and GPT-4 for several reasons. First, ChatGPT o1 represents the most recent publicly accessible model from OpenAI, providing a more equitable benchmark against our AD-GPT, which is similarly designed for practical deployment without proprietary access constraints. In contrast, models like GPT-4 operate behind closed APIs, limiting transparency, reproducibility, and direct evaluation. Furthermore, state-of-the-art LLMs such as GPT-4 frequently integrate online retrieval mechanisms to augment responses, whereas our model relies solely on internalized domain knowledge to ensure consistency and reliability in offline environments. The inclusion of retrieval-augmented models would introduce external variability, complicating direct comparisons. Finally, earlier versions such as GPT-3 exhibit insufficient specialization in genetic and biomedical domains, rendering them less relevant as comparators for our fine-tuned model.

% \begin{figure}[ht]
%     \centering
%     \includegraphics[width=0.7\linewidth]{gemini_response.png}
%     \caption{Gemini1.5-Flash response for task1}
%     \label{fig:10}
% \end{figure}

%\subsection{The Advantages}\label{subsec5}
%\subsubsection{Decisive answers and accurate references}\label{subsubsec2}
Based on our model, AD-GPT provided more decisive responses and produced more accurate references.
Our comparative analysis revealed that while state-of-the-art large language models (LLMs), such as ChatGPT-01, Claude 3.5-Haiku, and Gemini 1.5-Flash, demonstrated remarkable versatility across diverse tasks, they exhibited notable limitations in domain-specific expertise. These models, despite their extensive parameter sizes, relied heavily on web searches and external data sources to enhance reasoning accuracy and reference reliability. This dependence on online access introduced inconsistencies, particularly when handling specialized knowledge.
We observed that although the responses generated by these LLMs appeared logically structured and contextually appropriate, they often lacked decisiveness and contained ambiguities. More critically, when encountering gaps in knowledge, these models frequently produced fabricated references, i.e., citations that were either nonexistent or inaccurately attributed. This phenomenon led to the dissemination of misleading information, raising concerns about their reliability in research and decision-making contexts (Fig. \ref{fig:limitations}). 
In contrast, AD-GPT provided precise and well-reasoned responses derived exclusively from its curated domain-specific database. Unlike conventional LLMs, our model did not generate fictitious references, ensuring that all cited information remained verifiable and directly traceable. This fundamental advantage positioned AD-GPT as a more dependable tool for specialized knowledge retrieval, particularly in high-stakes applications where accuracy and authoritative sourcing were paramount.

We have achieved easy deployment and data transparency for AD-GPT. In contrast to large-scale LLMs, such as the ChatGPT series, our AD-GPT operated as a fully self-contained system with a compact architecture. This design facilitated secure, on-premise deployment, faster inference times, and lower latency, making it feasible for real-time applications in clinical and research settings. Furthermore, the knowledge base of our AD-GPT was constructed from a rigorously curated dataset specifically tailored to AD. Unlike models that aggregated information from broad-spectrum training corpora or relied on real-time web searches, our system ensured that all embedded knowledge was sourced from validated, authoritative references. This database-driven approach not only enhanced reliability but also enabled explicit citation of sources, a crucial feature for clinical decision support and research applications. By providing transparent, evidence-backed responses, our model aligned with the stringent requirements of medical and scientific fields, offering a trustworthy tool for domain experts.

% \begin{figure}[ht]
%     \centering
%     \includegraphics[width=0.75\linewidth]{fakereferenceo1.png}
%     \caption{Fake reference generated by Chatgpt o1}
%     \label{fig:11}
% \end{figure}

%\section{Outlook}
Our future efforts will focus on several key enhancements to improve the adaptability, precision, and scalability of our AD-GPT.
First, we plan to incorporate Retrieval-Augmented Generation (RAG) \cite{lewis2020retrieval} to address the limitations of our current fixed-database approach. While our model effectively answers domain-specific questions based on a predefined dataset, its knowledge remains constrained by the static nature of its training corpus. As research in AD evolves, maintaining up-to-date insights becomes increasingly critical. By integrating RAG, we aim to dynamically retrieve relevant information from external sources, such as recent publications, NCBI articles, and continuously updated medical databases. This enhancement will allow our model to generate more informed and contextually relevant responses, balancing domain-specific expertise with real-time knowledge integration. Furthermore, we intend to implement Chain-of-Thought (CoT) prompting \cite{wei2022chain} to improve logical coherence and reasoning depth in our responses. By enabling the model to generate intermediate reasoning steps, CoT enhances interpretability and ensures a structured decision-making process. The combination of RAG and CoT will not only refine response quality but also mitigate hallucinations, strengthening reliability in medical and research applications.
Second, inspired by DeepSeek R1 \cite{guo2025deepseek}, we plan to integrate Mixture-of-Experts (MoE) and reinforcement learning (RL) to further enhance response precision and adaptability. While our current BERT classifier routes questions to separate models, MoE presents a transformative approach to multi-task learning by dynamically assigning queries to the most relevant subset of expert networks. This selective activation reduces computational and storage overhead while improving contextual awareness, facilitating nuanced knowledge transfer across tasks. Unlike maintaining full models for each task, MoE employs a shared base network with lightweight expert modules, enabling a more efficient and scalable framework for complex, multi-dimensional queries. Additionally, we will incorporate Guided Reinforcement with Preference Optimization (GRPO) to refine our model’s decision-making capabilities. By leveraging continuous feedback mechanisms, RL will enhance adaptive learning, allowing our system to iteratively improve based on user interactions and expert evaluations. The synergy between MoE and RL will optimize resource allocation while maintaining high-performance standards in specialized medical reasoning tasks.

\section{Methods}

\subsection{Data and data collection}
\label{sec:data_collection}
Genetic datasets have expanded significantly in both scale and diversity, providing a robust foundation for advancing research in AD. Our study leveraged a core set of 144 experimentally validated seed genes identified in \cite{fang2021endophenotype}, each associated with AD pathogenesis (Supplementary Data). This initial gene set served as a fundamental reference point for subsequent analyses and validation. While these 144 genes provided a starting framework, additional candidate genes could be integrated in future investigations to enhance the breadth and robustness of our findings.

\begin{figure}[h]
    \centering
    \includegraphics[width=1\linewidth]{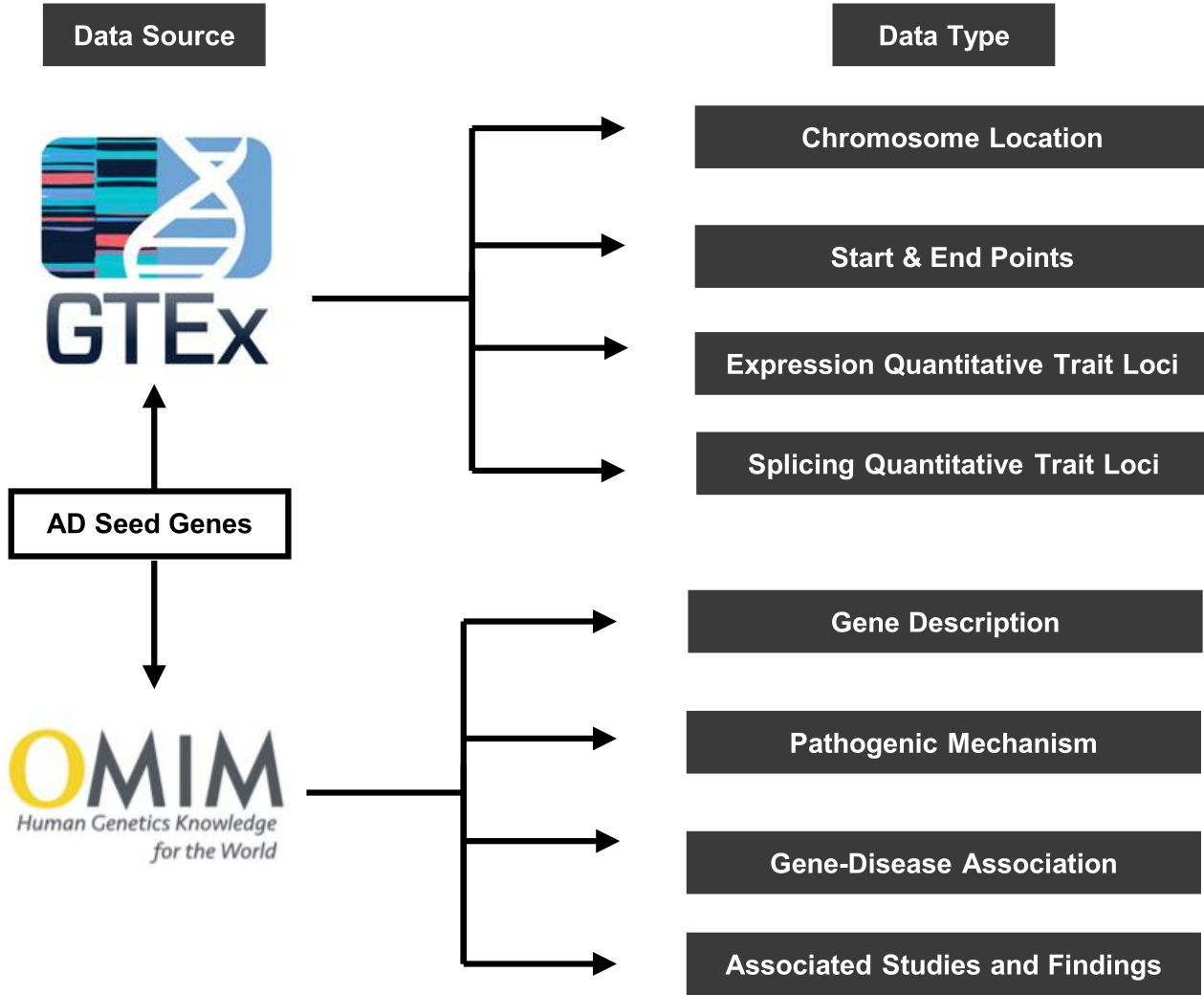}
    \caption{\textbf{Integration of genetic data from GTEx and OMIM. }GTEx provides gene location and QTL data, while OMIM offers gene descriptions and disease associations for AD research.}
    \label{fig:enter-label}
\end{figure}

Building on this foundation, we systematically curated and harmonized data from multiple high-quality sources, including the GTEx project and OMIM database (Fig. \ref{fig:enter-label}). The GTEx project provides extensive gene expression profiles across more than 30 non-diseased human tissue types collected from hundreds of donors. This resource enables the systematic exploration of associations between genetic variants identified in genome-wide association studies (GWAS) and disease phenotypes, offering critical insights into gene regulation mechanisms in AD. OMIM is a comprehensive and authoritative resource on human genes and genetic disorders. It contains detailed records on over 16,000 genes, including their functions, associated variants, molecular mechanisms, inheritance patterns, and links to relevant literature. This integrative approach ensured a comprehensive dataset, facilitating deeper insights into the genetic underpinnings of AD and enabling more precise downstream analyses.

From the GTEx database, we extracted cis-quantitative trait loci (cis-QTL) data, including both expression quantitative trait loci (eQTL) and splicing quantitative trait loci (sQTL), for the 144 AD seed genes using FastQTL \cite{ongen2016fast}, following the analysis pipeline in GTExPortal \cite{GTEx}
%eQTL analysis identifies genetic variants that modulate downstream gene expression, elucidating potential regulatory mechanisms underlying AD pathology. In parallel, sQTL analysis examines the impact of genetic variants on mRNA splicing, providing additional insights into transcriptomic alterations linked to AD. By integrating these complementary approaches, we aimed to uncover the functional consequences of AD-associated genetic variants, enhancing our understanding of the molecular basis of disease progression.
% To calculate cis-QTLs, gene expression and splicing values were quantified and mapped using FastQTL, with covariates including the top five genotype principal components, sequencing platform, sequencing protocol, sex, and additional hidden factors inferred using the Probabilistic Estimation of Expression Residuals (PEER) method \cite{stegle2010bayesian}. Nominal p-values were generated for each variant-gene pair by testing the alternative hypothesis that the slope of a linear regression model between genotype and expression deviates from zero. However, to control for multiple hypothesis testing and minimize the false discovery rate (FDR), Beta distribution-adjusted empirical p-values from FastQTL were used to calculate q-values, and a false discovery rate (FDR) threshold of $\leq 0.05$ was applied to identify genes with a significant eQTL \cite{goeman2014multiple}.
These data were obtained from a single tissue type encompassing 13 distinct brain regions, including the frontal cortex, amygdala, anterior cingulate cortex, caudate (basal ganglia), cerebellar hemisphere, cerebellum, nucleus accumbens (basal ganglia), putamen (basal ganglia), cervical spinal cord, cortex, hypothalamus, hippocampus, and substantia nigra.
Notably, gene-variant pairs often exhibit significant effects in one brain region but not in others, reflecting regional specificity in eQTL and sQTL distributions. These variations facilitate fine-mapping of potential causal variants and their corresponding neurological disease associations \cite{zhang2020regional}, providing a framework for identifying region-specific therapeutic targets for AD. Additionally, we extracted fundamental gene-level annotations, including chromosome location, start and end coordinates, and strand orientation, to support the genetic information retrieval. To further investigate the relationship between genotypes and AD, we further extracted the ``Molecular Genetic''  section from the OMIM entry of each gene (Supplementary Table 3),
 which provided a textual description about how genetic variants contribute to AD pathophysiology and established a robust basis for exploring genotype-phenotype relationships.

In summary, we curated a dataset of 144 AD-associated seed genes and integrated multi-omic regulatory information (Fig. \ref{fig:enter-label}). cis-eQTL, cis-sQTL, and basic gene information were obtained from the GTEx project, focusing on regional variations in gene expression and alternative splicing across 13 anatomically distinct human brain regions. This allowed us to construct gene-brain region associations and assess differential genetic functions within diverse neurological contexts. Additionally, we incorporated phenotypic and molecular data from the OMIM database to evaluate the functional consequences of these genes. By integrating these datasets, we established a comprehensive framework for elucidating the regulatory mechanisms underlying AD pathology and identifying potential therapeutic targets.

\subsection{Corpora construction}\label{sec:corpora}
To leverage the intrinsic relationships within multiple datasets, we constructed different corpora designed for multi-task, multi-level genomics fine-tuning. The corpora consists of four distinct training datasets, each corresponding to a specific task that encapsulates different aspects of genomics knowledge and context. Each corpus follows a standardized format aligning with instruction-based fine-tuning protocols for Llama models. Specifically, the corpora are structured with three key components: a system prompt, an input query, and a corresponding response. The system prompt is formulated as: ``\emph{You are a bioinformatics expert. Based on the following instruction, provide an accurate and professional response.}''. The input and response components vary across tasks to ensure adaptability and contextual relevance. 

For Task 1, AD-GPT is designed to provide precise genetic information by structuring gene-related data into a text-based format suitable for supervised fine-tuning. The dataset is constructed using gene symbols and their corresponding chromosomal locations, including start and end positions, derived from the GTEx database. To enhance the model’s adaptability, these data points are presented in multiple textual formats. A representative structured input query is: ``\emph{what is the start position of \{gene symbol\}?}". This formulation ensures that the model learns to retrieve and deliver accurate genomic coordinates in response to diverse query structures.

For Task 2, AD-GPT is designed to evaluate the relationship between specific genes and brain regions, with a particular emphasis on identifying significant genetic variations. To construct the corresponding training corpus, we compiled eQTL and sQTL data for seed genes across multiple brain regions, as curated from the GTEx portal. Gene-variant pairs were selected based on statistical significance, using q-values as the primary metric. A gene was considered associated with a brain region if it harbored one or more significant variants affecting expression or splicing functions in that region. This criterion enabled the establishment of robust gene–brain region links based on functional genomic evidence. The model was trained to generate binary responses to queries, ensuring precise and interpretable outputs. For instance, a representative query is: ``{\emph{Does the gene \{gene symbol\} contain variants in the \{brain region\} that significantly influence splicing regulation?}". This structured approach ensures the model delivers concise, evidence-based insights, facilitating efficient and reliable interpretation of gene–brain region interactions.

For Task 3, AD-GPT is designed to investigate the associations between specific genes and AD using molecular genetics information sourced from the OMIM database. Given that OMIM entries often contain extensive genetic details, not all of which are directly relevant to AD, we implemented a systematic curation process to extract meaningful relationships.
To refine the corpus, we manually identified and extracted the key reasoning components from the molecular genetics summaries, focusing on the most critical information that demonstrates gene-AD associations. This curated reasoning, along with the extracted dataset, was used as training material for the language model, enabling it to distinguish relevant genetic insights from broader molecular descriptions.
A typical query could be ``\emph{Determine if \{gene symbol\} has a potential role in Alzheimer's disease based on the molecular genetics summary.}'', and if the gene is related to AD, the answer will be ``\emph{Yes, there is a potential relation based on the \{reasoning\}}.'' 
During training, the model was optimized to generate structured responses comprising two essential components: (1) a definitive classification indicating whether a gene-AD relationship is supported and (2) a rationale explicitly outlining the molecular evidence that underpins this classification. By incorporating explicit reasoning into the training process, the model not only ensures evidence-based outputs but also enhances its capacity for inference in complex biomedical contexts.

In Task 4, AD-GPT is designed to model the tripartite relationships among genes, AD, and brain regions. Building upon the preceding datasets, this corpus employed a CoT prompting approach, which facilitates step-by-step reasoning to enhance performance on complex inferential tasks \cite{wei2022chain}. To illustrate, in response to the query, \emph{``Is the \{brain region\} related to AD with regard to gene \{gene symbol\}?''}, the reasoning process was explicitly structured into three sequential steps: (1) evaluating the association between the specified gene and AD, (2) assessing the relationship between the brain region and the gene, and (3) if both relationships were established, concluding that the brain region is related to AD with respect to the given gene. This structured reasoning process was embedded within the corpus, providing explicit guidance during training and improving the ability of AD-GPT in handling complex relational queries.

Representative examples of the four tasks are provided in Supplementary Tables 4–7, where each table illustrates a single example format per task. 
To enhance the model’s generalization capability, we incorporated diverse phrasings of the same query type within each corpus. For instance, in the task of determining a gene’s chromosomal location, training examples included variations such as ``\emph{What is the chromosome location of gene \{gene symbol\}?}'' and ``\emph{On which chromosome is \{gene symbol\} located?}''. This approach ensures robustness by enabling the model to recognize and accurately respond to different formulations of the same question.

\subsection{Competing LLMs for performance comparison}
\label{sec:LLM}
In this section, we summarize the key characteristics of competing LLMs that were used for performance comparison with AD-GPT.

ChatGPT o1 \cite{openai2024learning}, developed by OpenAI, represents a significant advancement in artificial intelligence reasoning capabilities. The model demonstrates exceptional performance across diverse domains, including competitive programming, advanced mathematics, and PhD-level scientific problem-solving. Notably, it exceeds human accuracy on standardized benchmarks in physics, biology, and chemistry, highlighting its potential for complex analytical tasks in biomedical research.

Claude3.5-Haiku \cite{anthropic_claude_3.5_haiku} is a language model developed by Anthropic, optimized for efficiency and cost-effectiveness while maintaining high performance across multiple skill sets. Despite its compact architecture, Claude3.5-Haiku demonstrates superior capabilities on various intelligence benchmarks, surpassing even Claude 3 Opus, the largest model of the previous generation, in key performance metrics. In February 2025, Anthropic introduced Claude3.7-Sonnet, the first hybrid reasoning model, which integrates enhanced contextual understanding, improved processing speed, and advanced problem-solving capabilities. This model represents a significant advancement in AI-driven analysis and decision-making, making it particularly well-suited for biomedical applications requiring nuanced interpretation of complex datasets.

Llama3.1-8B \cite{meta2025Llama31} is an open-source language model with 8 billion parameters, designed to enhance reasoning, computational efficiency, and comprehension of complex biomedical tasks. It features an extended context window and improved multilingual capabilities, making it well-suited for applications such as content generation, coding assistance, and biomedical text interpretation.
Llama3.2 was derived from Llama3.1 through a combination of structured pruning and knowledge distillation. Specifically, the 1B and 3B parameter models were obtained by systematically pruning Llama3.1-8B, reducing model complexity while preserving the original network’s performance. This optimization process aimed to enhance computational efficiency without compromising the model’s ability to process and generate high-quality biomedical text.

DeepSeek-R1-Distill-Llama-8B \cite{deepseek2025} is an open-source model developed by DeepSeek AI, derived from Llama3.1-8B through a distillation process. This distilled version maintains the high-performance capabilities of the original model while significantly improving computational efficiency. Benchmark evaluations have shown that DeepSeek-R1-Distill-Llama-8B achieves outstanding accuracy across a range of biomedical and general-domain tasks.

Gemini1.5 \cite{team2024gemini} is a model developed by Google DeepMind, which demonstrates near-perfect recall in long-context retrieval across multiple modalities. Gemini1.5 sets a new state-of-the-art performance in tasks such as long-document question answering (QA), long-video QA, and long-context automatic speech recognition (ASR). Additionally, we utilized Gemini1.5-Flash, a lightweight variant of the model, designed for enhanced efficiency while minimizing regression in performance quality.

Qwen2.5-7B \cite{qwen2.5} is a base model within the Qwen2.5 series of large language models, comprising 7.61 billion parameters. It utilizes a transformer architecture that incorporates advanced techniques such as Rotary Position Embedding (RoPE), SwiGLU activation, RMSNorm, and Attention QKV bias. The model supports a context window of up to 131,072 tokens, enabling it to process long-range dependencies effectively.

Grok3 \cite{xai2025grok3}, developed by xAI, is a state-of-the-art artificial intelligence model trained on the Colossus super-cluster, utilizing 10 times the computational power of previous leading models. This significant computational scaling enables Grok3 to achieve substantial improvements in tasks requiring reasoning, mathematics, coding, and instruction-following, facilitated by large-scale reinforcement learning. 

\subsection{Model architecture}\label{subsec15}
\begin{figure}
    \centering
    \includegraphics[width=\textwidth]{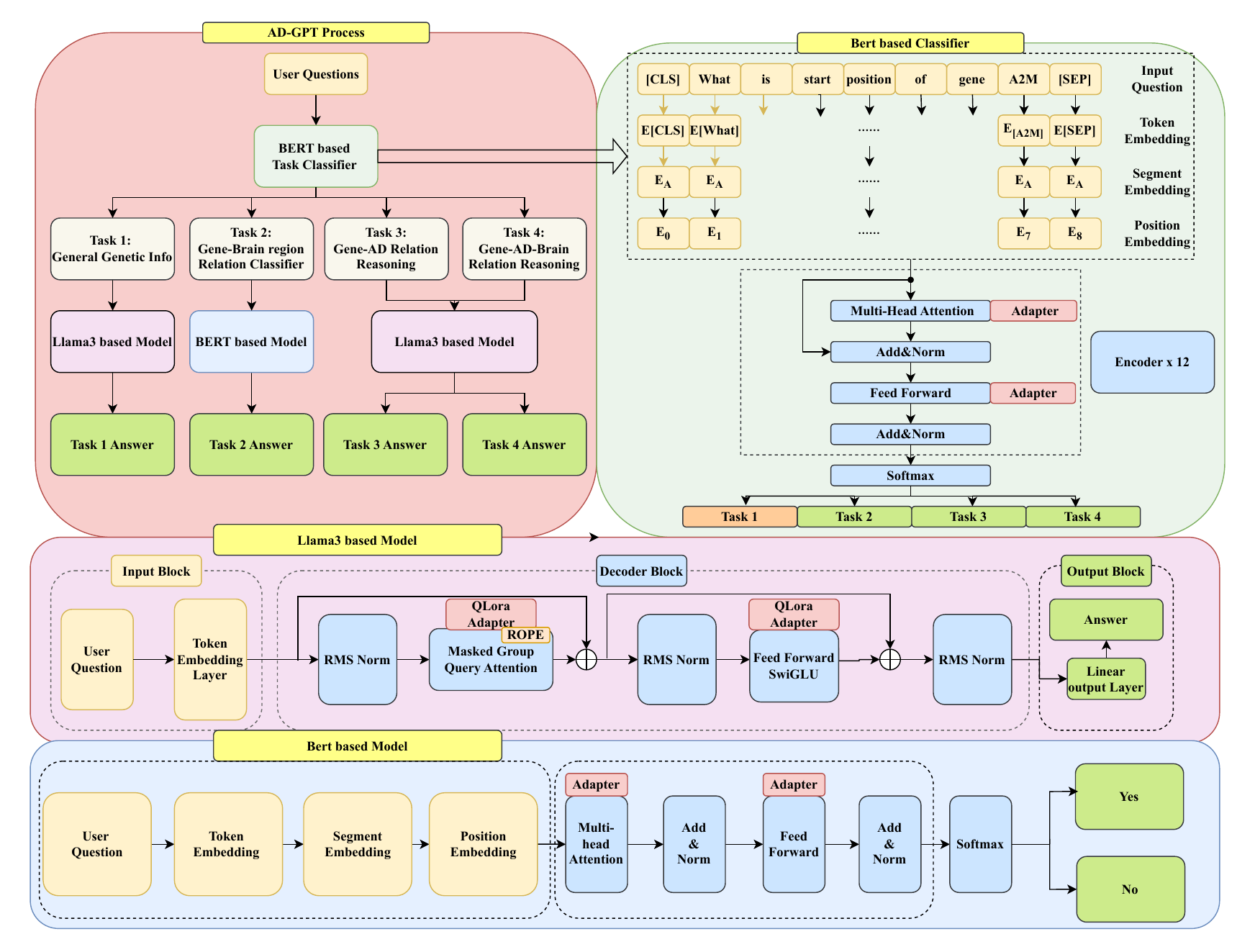}
    \caption{\textbf{AD-GPT architecture overview.} The AD-GPT model employs a hybrid architecture that integrates both BERT-based and Llama3-based models to address AD-related IR tasks. }
    \label{fig:Llamabert}
\end{figure}

 AD-GPT utilizes a classification-driven architecture to efficiently direct user queries to the most suitable expert model, as illustrated in Fig. \ref{fig:Llamabert}. At the core of this design is a classification model, BERT, which first identifies the category of the user’s query and subsequently routes it to the relevant expert model. This approach, inspired by the MoE paradigm, ensures that each query is processed by the model best equipped to handle it, thereby improving both accuracy and computational efficiency.

AD-GPT is comprised of the BERT model dedicated to Task 2 and the Llama3.1-8B model used for Tasks 1, 3, and 4. 
The BERT-based component is utilized for token-level classification tasks and incorporates token, segment, and position embeddings. Its encoder consists of 12 transformer layers featuring multi-head attention and feedforward networks, followed by normalization layers before task-specific classification heads. Fine-tuning is performed using adapter layers, enabling efficient transfer learning while preserving the pre-trained backbone. 
%This approach primarily updates the feedforward and classification layers, resulting in a computationally efficient model for gene-brain region inquiries. 
For reasoning-intensive tasks, the Llama3.1-8B model uses a decoder-only transformer architecture.
%, enhanced with Grouped-Query Attention (GQA) for efficient inference. 
Unlike BERT’s absolute positional embeddings, the Llama3.1-8B model utilizes rotary position embeddings, which encode relative positional information within the self-attention mechanism, improving the model’s ability to handle long-range dependencies. 
%Fine-tuning for the Llama3.1-8B model is carried out via QLoRA, which updates only the attention and feedforward layers while keeping the base model frozen. This method significantly reduces memory usage while maintaining strong performance in complex reasoning tasks.

\subsection{Fine tuning parameters}\label{subsec13}
QLoRA is an advanced fine-tuning technique designed to optimize LLMs with minimal computational and memory overhead. By utilizing 4-bit quantization, QLoRA significantly reduces the memory requirements of LLMs, enabling fine-tuning on resource-constrained hardware without compromising performance.
QLoRA was applied to fine-tune Llama3.1-8B models for Tasks 1, 3, and 4. Specifically, we fine-tuned the weights of the grouped-query attention (GQA) layer and the feedforward layer. For this process, we used a low-rank adaptation dimension ($r=64$), a scaling factor ($lora_{alpha}=16$), and a dropout rate ($dropout=0.1$) to balance computational efficiency with model generalization. This configuration allowed us to adapt the Llama3.1-8B models effectively to domain-specific tasks while leveraging QLoRA’s cost-effective and scalable framework to achieve high-quality results.
For Task 2, we fine-tuned the adapter layers within the feedforward and classification layers of a BERT-based model. This approach enabled task-specific adaptation while keeping the transformer backbone frozen, thereby preserving the model’s pre-trained linguistic knowledge while adapting it efficiently to biomedical tasks.

\subsection{Tokenization and embedding}\label{subsec14}
For both the Llama3.1-8B and BERT-based models, we retained the original tokenizers and embedding layers from the pre-trained versions, without introducing additional special tokens. While gene names and brain regions represent domain-specific terminology, we chose to preserve the pre-existing vocabulary and subword tokenization, thereby leveraging the models’ inherent linguistic and contextual knowledge.

The Llama3.1-8B model maintained its original embedding layer configuration of (128, 256, 4096), ensuring efficient representation learning without introducing additional computational overhead. Similarly, the BERT-based model retained its embedding layer structure of (30,522, 768), maintaining full compatibility with its pre-trained architecture.

Despite initial concerns that subword tokenization might limit domain-specific understanding, our fine-tuning experiments demonstrated that both models effectively captured complex gene-disease relationships, biological entity interactions, and mediation effects. This outcome was achieved without the need for vocabulary expansion. By preserving the original tokenizer and embedding layers, we optimized computational efficiency while ensuring strong performance on specialized biomedical tasks.

% \subsection{Graphical interface}\label{subsec16}
% Our graphical interface provides an intuitive way to interact with deployed models. All our models and FastAPI backend files are stored in a Docker container, making deployment as simple as starting the Docker instance. Once the container is running, we can easily use an HTML-based chat interface to communicate with the models. This setup ensures a smooth workflow, where the backend efficiently processes requests while the frontend delivers a seamless user experience. With this approach, we can quickly deploy, scale, and interact with our models in a structured and efficient manner.

%\subsection{Graphical interface}\label{subsec16}
%Our graphical interface provides an intuitive way to interact with deployed models. All our models and FastAPI backend files are stored in a Docker container, making deployment as simple as starting the Docker instance. Once the container is running, we can easily use an HTML-based chat interface to communicate with the models. This setup ensures a smooth workflow, where the backend efficiently processes requests while the frontend delivers a seamless user experience. With this approach, we can quickly deploy, scale, and interact with our models in a structured and efficient manner.

\newpage

\bibliography{sn-bibliography_cached}

\end{document}

\title[Article Title]{Supplementary information of AD-GPT: Large Language Models in Alzheimer's Disease}

\author[1]{\fnm{Ziyu} \sur{Liu}}\email{zl23565@uga.edu}

\author[2]{\fnm{Lintao} \sur{Tang}}\email{lt20ca@fsu.edu}

\author[3]{\fnm{Zeliang} \sur{Sun}}\email{zs58484@uga.edu}

\author[4]{\fnm{Zhengliang} \sur{Liu}}\email{zl18864@uga.edu}

% \author[5]{\fnm{Tianyang} \sur{Zhong}}\email{tzhong3@ualberta.ca}

\author[5]{\fnm{Yanjun} \sur{Lyu}}\email{yxl9168@mavs.uta.edu}

\author[4]{\fnm{Wei} \sur{Ruan}}\email{wei.ruan@uga.edu}

\author[2]{\fnm{Yangshuang} \sur{Xu}}\email{yx23e@fsu.edu}

\author[2]{\fnm{Liang} \sur{Shan}}\email{ls21s@fsu.edu}

\author[1]{\fnm{Jiyoon} \sur{Shin}}\email{js12571@uga.edu}

\author[2]{\fnm{Xiaohe} \sur{Chen}}\email{xchen8@fsu.edu}

\author[5]{\fnm{Dajiang} \sur{Zhu}}\email{dajiang.zhu@uta.edu}

\author[4]{\fnm{Tianming} \sur{Liu}}\email{tliu@uga.edu}

\author*[1]{\fnm{Rongjie} \sur{Liu}}\email{rjliu@uga.edu}

\author*[3]{\fnm{Chao} \sur{Huang}}\email{chaohuang@uga.edu}

\affil[1]{\orgdiv{Department of Statistics}, \orgname{University of Georgia}, \orgaddress{\city{Athens}, \state{GA}, \country{USA}}}

\affil[2]{\orgdiv{Department of Statistics}, \orgname{Florida State University}, \orgaddress{\city{Tallahassee}, \state{FL}, \country{USA}}}

\affil[3]{\orgdiv{Department of Epidemiology \& Biostatistics}, \orgname{University of Georgia}, \orgaddress{\city{Athens}, \state{GA}, \country{USA}}}

\affil[4]{\orgdiv{School of Computing}, \orgname{University of Georgia}, \orgaddress{\city{Athens}, \state{GA}, \country{USA}}}

% \affil[5]{\orgdiv{Department of Mathematical and Statistical Science}, \orgname{University of Alberta}, \orgaddress{\city{Edmonton},\country{Canada}}}

\affil[5]{\orgdiv{Department of Computer Science and Engineering}, \orgname{University of Texas at Arlington}, \orgaddress{\city{Arlington}, \state{TX}, \country{USA}}}


% \abstract{Large Language Models (LLMs) have proven to be valuable tools for information retrieval, serving as extensive reservoirs of knowledge in the medical field. The need for specialized information retrieval in Alzheimer's disease (AD) has become increasingly critical, as this condition poses an urgent societal challenge. However, concerns persist regarding the accuracy and depth of medical information provided by LLMs in Alzheimer's disease. To address this gap, we propose an Alzheimer's disease based Generative Pre-trained Transformer (AD-GPT), a novel tool designed specifically for Alzheimer’s disease research. AD-GPT integrates comprehensive data sources, including potential genes, molecular genetic information, significant gene variants in brain regions, and fundamental details about genes and brain regions associated with AD. In this paper, we develop a stacked LLM model architecture that incorporates Llama2 (7B), BERT to target four distinct tasks: (1) genetic information retrieval, (2) gene-brain region relationship assessment, (3) gene-Alzheimer’s disease relationship analysis and (4) brain region-Alzheimer’s disease relationship mapping. As conclusion, this paper outlines the framework and methodology for building AD-GPT, highlighting its potential to advance precision research and facilitate targeted discoveries in Alzheimer’s disease.}

\maketitle

\begin{table}[p]  % Place the table on a separate page
    \centering
    \renewcommand{\arraystretch}{1.1} % Adjust row spacing
    \setlength{\tabcolsep}{4pt} % Adjust column spacing
    \small % Reduce font size

    \caption*{\textbf{Supplementary Table 1: }\textbf{Performance comparison of models for Tasks 1 and 2.} 
    The metrics evaluate each model’s ability to answer gene attributed-related questions (Task 1) and evaluating gene-brain region relationships (Task 2) accurately and consistently.}
    \label{tab:task1_task2_performance}

    \begin{tabular}{lcccccc}
        \toprule
        \textbf{Model} & \textbf{Params} & \multicolumn{1}{c}{\textbf{Task 1 Acc. (\%)}} & \multicolumn{3}{c}{\textbf{Task 2 Metrics (\%)}} \\
        \cmidrule(lr){3-3} \cmidrule(lr){4-6}
        & & \textbf{Accuracy} & \textbf{Precision} & \textbf{Recall} & \textbf{F1 Score} \\
        \midrule
        Base AD-GPT  & 8B   & 90.84 & - & - & - \\
        Llama3.1   & 70B   & 70.52 & 67.52 & 95.82 & 79.22 \\
        ChatGPT o1  & $\approx$ 200B    & 85.33 & - & - & - \\
        Llama2 & 70B      & 59.59 & - & - & - \\
        Claude3.5-Haiku & $\approx$ 175B      & 74.28 & - & - & - \\
        Qwen2.5 & 7B & 17.14 & 99.83 & 67.33 & 80.42 \\
        Base AD-GPT (BERT)  & 110M   & - & 100.0 & 100.0 & 100.0 \\
        DeepSeek-Distilled-Llama-3.1 & 8B & - & 65.99 & 100.0 & 79.51 \\
        Llama3.2 & 1B & - & 70.41 & 95.43 & 81.03 \\
        \bottomrule
    \end{tabular}

    \vspace{3pt} % Small space between table and note
    \raggedright % Align notes to the left
    \textbf{Note:} Task 1 is evaluated using accuracy, while Task 2 is evaluated using accuracy, precision, recall, and F1-score. Values are in percentages.

\end{table}

\begin{landscape}
\begin{table}[p]  % Place the table on a separate page
    \centering
    \renewcommand{\arraystretch}{1.1} % Adjust row spacing
    \setlength{\tabcolsep}{6pt} % Adjust column spacing
    \small % Reduce font size

    \caption*{\textbf{Supplementary Table 2: }\textbf{Performance comparison of models for Task 3 and Task 4.} 
    The metrics evaluate each model’s precision and relevance in different tasks.}
    \label{tab:task3_task4_performance}

    \begin{tabular}{lcccc}
        \toprule
        \textbf{Model} & \multicolumn{2}{c}{\textbf{Task 3 Metrics}} & \multicolumn{2}{c}{\textbf{Task 4 Metrics}} \\
        \cmidrule(lr){2-3} \cmidrule(lr){4-5}
        & \textbf{Avg Precision $\pm$ SD} & \textbf{Avg Relevance $\pm$ SD} & \textbf{Avg Precision $\pm$ SD} & \textbf{Avg Relevance $\pm$ SD} \\
\midrule
AD-GPT         & 4.70 $\pm$ 0.73 & 4.92 $\pm$ 0.24 & 4.60 $\pm$ 0.90 & 4.90 $\pm$ 0.27 \\
ChatGPT o1              & 3.88 $\pm$ 1.10 & 4.52 $\pm$ 0.51 & 3.67 $\pm$ 0.82 & 4.62 $\pm$ 0.29 \\
Gemini1.5-Flash       & 2.88 $\pm$ 1.89 & 4.75 $\pm$ 0.42 & 3.63 $\pm$ 0.69 & 4.57 $\pm$ 0.31 \\
Claude3.5-Haiku       & 3.35 $\pm$ 1.24 & 4.37 $\pm$ 0.55 & 3.87 $\pm$ 0.72 & 4.77 $\pm$ 0.34 \\
Deepseek-R1-Distill-Llama       & 3.23 $\pm$ 0.43 & 4.63 $\pm$ 0.28 & 3.30 $\pm$ 0.42 & 4.33 $\pm$ 0.26 \\
Llama3.1-8B    & 3.67 $\pm$ 0.69 & 4.78 $\pm$ 0.25 & 3.25 $\pm$ 0.53 & 4.28 $\pm$ 0.31 \\
Claude3.7-Sonne       & 4.45 $\pm$ 0.47 & 4.82 $\pm$ 0.20 & 4.33 $\pm$ 0.36 & 4.37 $\pm$ 0.42 \\
Grok3         & 4.13 $\pm$ 0.50 & 4.62 $\pm$ 0.33 & 4.22 $\pm$ 0.44 & 4.62 $\pm$ 0.33 \\
Llama3.2-3B       & 3.65 $\pm$ 0.44 & 4.80 $\pm$ 0.27 & 3.23 $\pm$ 0.36 & 4.20 $\pm$ 0.33 \\
\bottomrule
    \end{tabular}

    \vspace{3pt} % Small space between table and note
    \raggedright % Align notes to the left
    \textbf{Note:} Task 3 and Task 4 are evaluated using average precision (0-5) and average relevance scores (0-5) rated by experts.
    
\end{table}
\end{landscape}

\begin{table}[h]
\caption*{\textbf{Supplementary Table 3: }Example of molecular genetics content extracted from the OMIM database. This shows partial raw text from the APOE gene description, while red text presents manually extracted reasoning, highlighting key information relevant to the relationship between APOE and Alzheimer’s disease.}\label{reason}
\begin{tabular}{@{}p{12cm}@{}}
\toprule
\textbf{Partial content in molecular genetic description of APOE on OMIM website}   \\
\midrule
Data on gene frequencies of apoE allelic variants were tabulated by Roychoudhury and Nei (1988).
In a comprehensive review of apoE variants, de Knijff et al. (1994) found that 30 variants had been characterized, including the most common variant, apoE3.\\
...\\
Hyperlipoproteinemia Type III\\
Smit et al. (1987) described 3 out of 41 Dutch dysbetalipoproteinemic patients who were apparent E3/E2 heterozygotes rather than the usual E2/E2 homozygotes. All 3 genetically unrelated patients showed an uncommon E2 allele that contained only 1 cysteine residue. The uncommon allele cosegregated with familial dysbetalipoproteinemia which in these families seemed to behave as a dominant. Smit et al. (1990) showed that these 3 unrelated patients had (E2K146Q; 107741.0011).
\\
...\\
Alzheimer Disease 2\\
{\color{red}Saunders et al. (1993) reported an increased frequency of the E4 allele in a small prospective series of possible-probable Alzheimer disease patients presenting to the memory disorders clinic at Duke University, in comparison with spouse controls. Corder et al. (1993) found that the APOE*E4 allele is associated with the late-onset familial and sporadic forms of Alzheimer disease. In 42 families with the late-onset form of Alzheimer disease (AD2;1 04310), the gene had been mapped to the same region of chromosome 19 as the APOE gene. {\color{red}Corder et al. (1993) found that the risk for AD increased from 20 to 90\% and mean age of onset decreased from 84 to 68 years with increasing number of APOE*E4 alleles. Homozygosity for APOE*E4 was virtually sufficient to cause AD by age 80.}}

\\

\bottomrule
\end{tabular}
\end{table}

\begin{table}[h]
\caption*{\textbf{Supplementary Table 4: }Task 1 consists of three sub-tasks related to chromosome location, start position, and end position. The data, sourced from the GTEx project, are structured into a textual corpus for model training. The blue text represents user-provided key information, while the red text indicates the key responses generated by the LLM model}\label{tab:task1}
\begin{tabular}{@{}p{5cm}p{7cm}@{}}
\toprule
\textbf{System Prompt}  & You are a bioinformatics expert, Based on the following instruction, provide an accurate and professional response \\
\midrule
\textbf{Instruction Formats}  & What is the chromosome location of the gene 
{\color{blue}\textit{\{gene symbol\}}} on corresponding chromosome\\& What is the start position of 
{\color{blue}\textit{\{gene symbol\}}}?\\& What is the end position of 
{\color{blue}\textit{\{gene symbol\}}}?
 \\
\midrule
\textbf{Output Formats} & The chromosome location of gene {\color{blue}\textit{\{gene symbol\}}} is {\color{red}\textit{\{chromosome\}}}.\\&The start position of {\color{blue}\textit{\{gene symbol\}}} is {\color{red}\textit{\{start position\}}} on {\color{blue}\textit{\{chromosome\}}}.\\& The end position of {\color{blue}\textit{\{gene symbol\}}} is {\color{red}\textit{\{end position\}}} on {\color{blue}\textit{\{chromosome\}}}.
 \\
\midrule
\textbf{Model for Fine-tuned } & Llama3.1-8B \\
\textbf{Included Instruction-output pairs} & $2160$\\
\textbf{Data Source} & GTEx\\

\bottomrule
\end{tabular}
\end{table}

\begin{table}[h]
\caption*{\textbf{Supplementary Table 5: }Task 2 formulates the relationship between brain regions and genes as a binary ``Yes or No'' question. The relationships, derived from eQTL and sQTL data, are transformed into a structured textual corpus. The blue text represents user-provided key information, while the red text indicates the key responses generated by the LLM model }\label{tab:task2}
\begin{tabular}{@{}p{5cm}p{7cm}@{}}
\toprule
\textbf{System Prompt}  & You are a bioinformatics expert, Based on the following instruction, provide an accurate and professional response \\
\midrule
\textbf{Instruction Formats}  & Does the gene {\color{blue}\textit{\{gene symbol\}}} have any significant variants in the {\color{blue}\textit{\{brain region\}}} region that significantly affect the expression regulation? \\& Does the gene {\color{blue}\textit{\{gene symbol\}}} have any significant variants in the {\color{blue}\textit{\{brain region\}}} region that significantly affect the splicing regulation?
\\
\midrule
\textbf{Output Format} & {\color{red}\textit{\{Yes\}}} \textbf{or} {\color{red}\textit{\{No\}}}. \\
\\
\midrule
\textbf{Model for Fine-tuned } & BERT \\
\textbf{Included Instruction-output pairs} & $10140$\\
\textbf{Data Source} & GTEx\\

\bottomrule
\end{tabular}
\end{table}

\begin{table}[h]
\caption*{\textbf{Supplementary Table 6: }Task 3 focuses on determining whether a specific gene has a potential role in Alzheimer’s disease based on its molecular genetics summary. The molecular genetics summary, sourced from the OMIM database, serves as the foundation for the input prompt. The model evaluates this information and provides a ``Yes'' or ``No'' response, along with reasoning based on the OMIM data. The blue text represents user-provided key information, while the red text indicates the key responses generated by the LLM model}\label{tab:task3}
\begin{tabular}{@{}p{5cm}p{7cm}@{}}
\toprule
\textbf{System Prompt}  & You are a bioinformatics expert, Based on the following instruction, provide an accurate and professional response \\
\midrule
\textbf{Instruction Formats}  & Determine if {\color{blue}\textit{\{gene symbol\}}} has a potential role in Alzheimer's disease based on the molecular genetics summary.
\\
\midrule
\textbf{Output Formats} & {\color{red}\textit{\{Yes\}}}, there is potential relation based on {\color{red}\textit{\{reasoning\}}}.\\&\textbf{or} {\color{red}\textit{\{No\}}}, there is no potential relation found based on our database.
 \\
\\
\midrule
\textbf{Model for Fine-tuned } & Llama3.1-8B \\
\textbf{Included Instruction-output pairs} & $3040$\\
\textbf{Data Source} & OMIM\\

\bottomrule
\end{tabular}
\end{table}

\begin{table}[h]
\caption*{\textbf{Supplementary Table 7: }Example of Task 4 prompt and response format. The prompt queries the relationship between a brain region and AD concerning a specific gene, while the response provides an evidence-based assessment of the gene-AD and gene-brain region relationships to conclude the potential brain region-AD association. The blue text represents user-provided key information, while the red text indicates the key responses generated by the LLM model}\label{tab:task4}
\begin{tabular}{@{}p{5cm}p{7cm}@{}}
\toprule
\textbf{System Prompt}  & You are a bioinformatics expert, Based on the following instruction, provide an accurate and professional response \\
\midrule
\textbf{Instruction Formats}  & Is {\color{blue}\textit{\{brain region\}}} related to AD regarding the gene {\color{blue}\textit{\{gene symbol\}}}?

\\
\midrule
\textbf{Output Formats} & Based on the molecular genetic summary of {\color{blue}\textit{\{gene symbol\}}} there {\color{red}\textit{\{is\}}} \textbf{or} {\color{red}\textit{\{is not\}}}
 potential relationship between {\color{blue}\textit{\{gene symbol\}}} and AD. And {\color{blue}\textit{\{gene symbol\}}} {\color{red}\textit{\{has\}}} \textbf{or} {\color{red}\textit{\{does not have\}}} significant variants on region {\color{blue}\textit{\{brain region\}}}, so there {\color{red}\textit{\{is\}}} \textbf{or} {\color{red}\textit{\{is not\}}} potential relationship between {brain region} and AD.

 \\
\\
\midrule
\textbf{Model for Fine-tuned } & Llama3.1-8B \\
\textbf{Included Instruction-output pairs} & $11115$\\
\textbf{Data Source} & OMIM, GTEx\\

\bottomrule
\end{tabular}
\end{table}